\documentclass{article}

\usepackage[preprint]{neurips_2026}

\usepackage[utf8]{inputenc} 
\usepackage[T1]{fontenc}    
\usepackage{hyperref}       
\usepackage{url}            
\usepackage{booktabs}       
\usepackage{amsfonts}       
\usepackage{amsmath}
\usepackage{amssymb}
\usepackage{nicefrac}       
\usepackage{microtype}      
\usepackage{xcolor}         
\usepackage{graphicx}
\usepackage{cleveref}
\usepackage{bm}
\usepackage{float}
\usepackage{placeins}
\usepackage{multirow}

\title{Scalable Bayesian Gaussian Process Classification for Cancer Histopathology: Calibrated Uncertainty Under Distribution Shift}

\author{%
  Yunru Zheng \quad Wun Ting Chan \quad Sampras Dsouza \quad Zhenyu Hong \\
  New York University \\
  \texttt{\{yz9076,wc3013,zh2746,sd6701\}@nyu.edu}
}

\begin{document}

\maketitle

\begin{abstract}
    Reliable uncertainty quantification is essential for clinical deployment of cancer diagnosis models, yet neural networks and variational Gaussian process methods can produce overconfident predictions, particularly under the distribution shifts that arise when models trained at one hospital are applied at another. We present a scalable Bayesian Gaussian Process Classification (GPC) framework that operates on frozen pathology foundation model embeddings (\textit{Phikon}) and combines accelerated Randomly Pivoted Cholesky (RPCholesky) for low-rank kernel factorization, Hamiltonian Monte Carlo (HMC) with non-centered parameterization for posterior sampling, and a pivots-as-inducing-points scheme for prediction. This makes HMC posterior sampling in the low-rank latent representation tractable at $N\approx300,000$ training points, enabling HC-based low-rank GPC on substantially larger datasets than prior MCMC-based demonstrations. We evaluate on two histopathology benchmarks for breast cancer metastasis detection: \textbf{PCam} and \textbf{CAMELYON17-WILDS}, where the latter tests calibration under hospital-level distribution shift. Compared to stochastic variational GP and neural network baselines, our method improves calibration and predictive performance — reducing expected calibration error by up to 88\% and false negative rate by up to 90\% — with the largest gains appearing under distribution shift, suggesting that HMC-based posterior sampling provides more trustworthy uncertainty estimates for cross-site pathology classification.
\end{abstract}

\section{Introduction}

Neural networks have been widely used for cancer diagnosis 
from histopathology images. Despite their high accuracy, neural networks can produce overconfident predictions\citep{guo2017calibration}, which is a serious limitation in clinical environments where uncertainty quantification is critical. For cancer histopathology, false negatives can be life threatening and delay treatment, whereas false positives can cause patient distress and unnecessary medical expenses. This problem is further compounded when models trained at one hospital are deployed at another, where differences in tissue preparation, staining protocols, and scanner hardware introduce distribution shifts that can silently degrade both accuracy and calibration in neural networks. As pathology foundation models like \textit{Phikon} are increasingly adopted across institutions \citep{filiot2023phikon}, the need for uncertainty-aware downstream classifiers that remain calibrated across hospital sites becomes critical for safe deployment.

In contrast, Gaussian Process Classification (GPC) provides principled uncertainty estimates. Unfortunately, the posterior has no closed form since the Bernoulli likelihood is non-conjugate with the Gaussian prior. Historically, approximate inference methods like stochastic variational Gaussian process (SVGP) have dominated GPC instead of direct sampling methods like MCMC \citep{nickisch2008approximations} due to their much lower cost. In this paper, we show that with a new kernel factorization method called accelerated randomly pivoted Cholesky (RPCholesky) \citep{epperly2025embrace}, we could use HMC, a type of Markov Chain Monte Carlo (MCMC) method to sample the posterior of GPC, which outperforms the SVGP and neural-network baselines on calibration and predictive metrics.

\paragraph{Contributions.} We make the following contributions: (1) We propose a scalable Bayesian GPC that combines accelerated RPCholesky \citep{epperly2025embrace} with HMC under non-centered parameterization, making MCMC-based GPC tractable at $N \approx 300{,}000$ training points; (2) We introduce a pivots-as-inducing-points predictive scheme based on previous literature (\citet{quinonero2005unifying}, \citet{williams2001nystrom}) that aligns the predictive inducing set with the RPCholesky training approximation; and (3) We demonstrate through cross-hospital evaluation on \textbf{CAMELYON17-WILDS} that HMC-based posterior sampling achieves substantially better calibration than SVGP and neural baselines on the held-out \textbf{CAMELYON17-WILDS} hospital, reducing ECE by up to 88\% and false negatives by up to 90\%.

\section{Related Work}
\paragraph{Low-rank kernel approximation.} 
Our method is directly inspired by the recently developed low rank approximate factorization algorithm for positive semi-definite matrices. \citet{chen2025randomly} introduce randomly pivoted Cholesky (RPCholesky), a variant of partial Cholesky decomposition that approximates positive semi-definite matrices with small trace-norm approximation error. \citet{epperly2025embrace} 
develop accelerated RPCholesky, which achieves up to 40 times faster factorization in practice via a combination of block RPCholesky and rejection sampling.

\paragraph{Approximate Inference for GP Classification and Stochastic variational Gaussian processes}
A long line of work on approximate inference addresses the intractable GPC posterior issue, surveyed by \citet{nickisch2008approximations}. Among different posterior computation methods, MCMC stands out with the highest posterior fidelity but has been considered intractable for large datasets. SVGP has emerged as the dominant scalable approach to GP inference with non-Gaussian likelihoods, which combines the inducing-variable framework of sparse GPs \citep{quinonero2005unifying,titsias2009variational}
with stochastic variational inference \citep{hensman2013gaussian}. \citet{hensman2015scalable} extend it to classification by optimizing a variational lower bound over a tractable Gaussian approximate posterior. However, variational inference based on the reverse
KL objective can underestimate posterior variance \citep{blei2017variational,yao2018yes}.
This loss of posterior uncertainty is especially concerning in safety-critical
classification problems, where miscalibrated probabilities and overconfidence
under distribution shift can have direct downstream clinical consequences
\citep{guo2017calibration,ovadia2019trust}. We
use SVGP classification \citep{hensman2015scalable} as our primary scalable GP
baseline. Recently, \citet{cooper2025modernizing} revived fully Bayesian GPC using elliptical slice sampling with Vecchia approximation, scaling to $n=10,000$, whereas we make GPC tractable at $N\approx 300,000$ training points, enabling MCMC-based low-rank GPC on datasets approximately 30× larger than the $n=10,000$ setting considered by \citet{cooper2025modernizing}.

\paragraph{Neural Networks} Deep learning has become the dominant approach for histopathology classification. Pathology foundation models such as \textit{Phikon} \citep{filiot2023phikon} learn transferable representations from large-scale self-supervised pretraining, and are typically paired with lightweight classification heads such as MLPs. Convolutional architectures like U-Net \citep{ronneberger2015unet} remain widely used in medical imaging, while Vision Transformers \citep{dosovitskiy2021vit} have shown competitive performance on image classification tasks with large-scale pretraining. These architectures achieve strong discriminative performance but lack built-in uncertainty quantification, typically requiring post-hoc calibration \citep{guo2017calibration}. We use all of them as baselines to our proposed method. 

\section{Method}
\subsection{Gaussian Process Classification}
Gaussian Process Classification (GPC) is a hierarchical Bayesian model \citep{rasmussen2006gaussian,nickisch2008approximations} with three levels: A kernel hyperparameter prior, a latent function with a GP prior, and the Bernoulli mapping that maps that latent function to binary labels. There are $N$ training points with dimension $d$ ($\mathbf{X}\in \mathbb{R}^{N\times d}$). We express the model in math language as follows.
\begin{enumerate}
    \item Hyperparameters: $\theta$: In general, the hyperparameter is $\theta = (s_f,l)$ for the kernel matrix with prior $p(\theta)$. The matrix entries are calculated using the kernel function. For any two data points $\mathbf{x}, \mathbf{x}'\in \mathbb{R}^d$ from the training points, the kernel function is   
    \begin{equation}\label{KernelFunction}
    k(\mathbf{x}, \mathbf{x}') = s_{f}^2\exp(-\frac{\|\mathbf{x} - \mathbf{x}'\|^2}{2l^2}).        
    \end{equation}
 The kernel matrix $\mathbf{K}_{\theta}$ is thus $N\times N$ and positive semi-definite \citep{rasmussen2006gaussian}. In our case, we fix $\theta$ while sampling the latent function $\mathbf{f}$, and do parameter sweeps later to find a good $\theta$.
    \item Latent function: $\mathbf{f}|\theta, \mathbf{X}\sim \mathcal{N} (0, \mathbf{K}_{\theta}(\mathbf{X},\mathbf{X}))$;
    \item Bernoulli mapping: $y_i|f_i \sim \text{Bernoulli} (\sigma(f_i))$. The mapping from latent function to binary labels is independent from $\theta$. We choose to use sigmoid function as the link function $\sigma$. 
\end{enumerate}

\subsection{Inference Problem}
When $\mathbf{f},\theta$ are jointly sampled, the posterior is 
\begin{equation} \label{posterior1}
  p(\mathbf{f},\theta|\mathbf{y},\mathbf{X}) = \frac{p(\mathbf{y}|\mathbf{f})\cdot p(\mathbf{f}|\theta,\mathbf{X})\cdot p(\theta)}{p(\mathbf{y}|\mathbf{X})}
\end{equation}

In our work, we fix the kernel amplitude at $s_{f}^2 = 1$, matching the kernel parameterization used in our accelerated RPCholesky implementation, and tune only the lengthscale. Jointly sampling $l$ with the latent function would require recomputing the low-rank factorization at each MCMC step, which is prohibitive at $N \approx 300,000$. We therefore fix $l$ ($\theta = (1,l)$) via parameter sweep (Section 4.3) and note that conditional on $l$, HMC samples the posterior over $f$ without further approximation beyond the low-rank kernel factorization. The kernel matrix can be referred to simply as $\mathbf{K}$, and the posterior we are sampling is 
\begin{equation} \label{posterior2}
p(\mathbf{f}|\mathbf{y},\mathbf{X},\theta) = \frac{p(\mathbf{y}|\mathbf{f})p(\mathbf{f}|\theta,\mathbf{X})}{p(\mathbf{y}|\theta,\mathbf{X})}    
\end{equation}
 The posterior of the latent function $p(\mathbf{f}|\mathbf{y},\mathbf{X},\theta)$ is generally not tractable since it is proportional to a Bernoulli likelihood multiplied by a Gaussian prior. In addition, the kernel matrix $\mathbf{K}$ is an $N\times N$ matrix, which costs $O(N^2)$ memory to store, and $O(N^3)$ time to factorize. Even for the smaller of our two training sets, \textbf{PCam}, $N = 262,144$, the exact matrix factorization is not possible.

\subsection{Low-rank Factorization: Accelerated RPCholesky}
The accelerated RPCholesky algorithm exploits the rapid spectral decay typical of kernel matrices \citep{chen2025randomly, epperly2025embrace} to factor the dense $N\times N$ kernel matrix $\mathbf{K}$ into low rank matrices:
\[
\mathbf{K} \approx \mathbf{L}\mathbf{L}^T\] for $\mathbf{L}\in \mathbb{R}^{N\times k}$. Using the AbstractPSDMatrix class defined in \citet{epperly2025embrace}, the low-rank factor is computed directly from the training data, while the actual $N\times N$ kernel matrix is never computed or stored. The input is an $N\times d$ matrix and the output is a $N\times k$ matrix $\mathbf{L}$ with low rank $k$, which is usually $O(10^2)$.

\subsection{Posterior Sampling: Hamiltonian Monte Carlo (HMC) with non-centered parameterization (NCP)}
    As mentioned above, the posterior of GPC $p(\mathbf{f}|\mathbf{y}, \mathbf{X}, \theta)$ is not tractable. SVGP addresses this issue by approximating the posterior with a Gaussian and optimizing via stochastic variational inference. An alternative for sampling this intractable posterior would be MCMC samplers, in particular HMC \citep{neal2011mcmc}. Our design using HMC for GPC is heavily inspired by \citet{betancourt2013hamiltonian}, which discuss using HMC for hierarchical models extensively.
    
    Naive HMC sampling of the posterior based on $\mathbf{f}\sim \mathcal{N}(0,\mathbf{K}_{\theta})$ is inefficient because the complicated kernel matrix induces strong correlation between latent values of nearby training points, creating elongated, and ill-conditioned posterior contours. We can use non-centered parameterization to solve this problem as follows: Set $\bm{\nu}\sim \mathcal{N}(0,I)$ and thus 
    \begin{equation}\label{eq:NCP}
    \mathbf{f} = \mathbf{L}_{\theta}\bm{\nu},    
    \end{equation}
    where $\bm{\nu}\in \mathbb{R}^k$ is the HMC-sampled latent function after non-centered parameterization, and $\mathbf{f} = \mathbf{L}_{\theta}\bm{\nu}$ is the actual training latent function. We now have an isotropic prior geometry, no longer sampling from $\mathbf{f}\sim \mathcal{N}(0, \mathbf{K}_{\theta})$ which has possible extreme eigenvalue spread. After NCP, we use Euclidean HMC as \citet{betancourt2013hamiltonian} suggest. This means that for $\rho\in \mathbb{R}^k$ the auxiliary momentum and $K(\rho)$ the kinetic energy in HMC,
    \begin{equation*}
        \rho\sim \mathcal{N}(0,\mathbf{I}), K(\rho) = \frac{1}{2}\rho^T\rho.
    \end{equation*}
    \citet{betancourt2013hamiltonian} give us the theoretical foundation for using HMC as the MCMC sampler for GPC, but in experiments we also compare the performance of different MCMC samplers on GPC at small scale to compare the relative sampling efficiency of the MCMC methods, before we use HMC on the full-scale dataset.

\subsection{Pivots as Inducing Points}
The posterior $p(\mathbf{f}|\mathbf{y}, \mathbf{X},\theta)$ describes the belief we have on training data, while the predictive distribution $p(\mathbf{f}_*|\mathbf{y}, \mathbf{X}_*)$ is the distribution over new  $n_{test}$ test points ($\mathbf{X}_*\in \mathbb{R}^{n_{test}\times d}$), assuming that the training and testing points are jointly Gaussian. Computing the predictive distribution can also be challenging at scale. For the joint Gaussian prior over training and testing latent values, \(\mathbf f\) and
\(\mathbf f_*\), a more accurate GP prediction requires solving linear systems involving
the \(N \times N\) training kernel matrix. This costs \(O(N^3)\) time for a dense
factorization and \(O(N^2)\) memory, which is infeasible for large \(N\). We
therefore use the RPCholesky pivots as inducing points, adapting ideas from
sparse variational GP approximations.

We adapt the inducing-variable method, specifically the deterministic training
conditional (DTC) approximation \citep{quinonero2005unifying}. Let
\(\mathcal P \subset \{1,\ldots,N\}\) denote the set of \(k\) RPCholesky pivot
indices, and let
\[
    \mathbf u = \mathbf f_{\mathcal P} \in \mathbb R^k
\]
be the latent function values at these pivot points. Under the DTC conditional
independence assumption
\[
    \mathbf f_* \perp \mathbf f \mid \mathbf u,
\]
prediction depends on the training data only through the \(k\)-dimensional
inducing variable \(\mathbf u\), reducing the dominant predictive linear algebra
from \(N\)-dimensional to \(k\)-dimensional computations.

Unlike standard inducing-variable methods that optimize inducing locations
during training \citep{snelson2005sparse,titsias2009variational}, we fix the
inducing points to be the RPCholesky pivots \(\mathcal P\). This choice is
natural in our setting because the pivots are already selected by the low-rank
kernel approximation procedure and correspond to informative columns of the
kernel matrix \citep{chen2025randomly}. The corresponding inducing variables
satisfy
\[
    \mathbf u = \mathbf f_{\mathcal P}
    \sim \mathcal N(0,\mathbf K_{\mathcal P,\mathcal P}).
\]
Moreover, the RPCholesky factor \(\mathbf L \in \mathbb R^{N \times k}\)
satisfies
\begin{equation}\label{PivotMatrixApprox}
    \mathbf L[\mathcal P,:]\mathbf L[\mathcal P,:]^T
    =
    \mathbf K_{\mathcal P,\mathcal P},
\end{equation}
so \(\mathbf K_{\mathcal P,\mathcal P}\) is available directly from the
precomputed low-rank factor. Therefore, for an HMC sample
\(\bm \nu\) with \(\mathbf f = \mathbf L \bm \nu\), the corresponding inducing
variable is
\[
    \mathbf u
    =
    \mathbf f_{\mathcal P}
    =
    \mathbf L[\mathcal P,:]\bm \nu .
\]

For a test set \(\mathbf X_*\), the DTC predictive conditional is
\begin{equation}\label{TestLatent}
    p(\mathbf f_* \mid \mathbf f_{\mathcal P}, \mathbf X_*)
    =
    \mathcal N \left(
        \mathbf K_{*,\mathcal P}
        \mathbf K_{\mathcal P,\mathcal P}^{-1}
        \mathbf f_{\mathcal P},
        \mathbf K_{*,*}
        -
        \mathbf K_{*,\mathcal P}
        \mathbf K_{\mathcal P,\mathcal P}^{-1}
        \mathbf K_{\mathcal P,*}
    \right).
\end{equation}
Here,
\(\mathbf K_{*,\mathcal P}=k(\mathbf X_*,\mathbf X_{\mathcal P})\) is the
test--pivot cross-kernel matrix,
\(\mathbf K_{\mathcal P,*}=\mathbf K_{*,\mathcal P}^T\), and
\(\mathbf K_{*,*}=k(\mathbf X_*,\mathbf X_*)\) is the test--test kernel matrix.
In practice, for marginal class probabilities we only need the diagonal of the
conditional covariance, so \(\mathbf K_{*,*}\) can be replaced by its diagonal.

The resulting predictive cost is
\[
    O(k^3 + n_{\mathrm{test}}k^2 + n_{\mathrm{test}}kd),
\]
where the \(O(k^3)\) term comes from factorizing
\(\mathbf K_{\mathcal P,\mathcal P}\), the \(O(n_{\mathrm{test}}k^2)\) term
comes from applying \(\mathbf K_{\mathcal P,\mathcal P}^{-1}\) to the
test--pivot kernel matrix, and the \(O(n_{\mathrm{test}}kd)\) term comes from
forming \(\mathbf K_{*,\mathcal P}\). Since in our experiments
\(k=O(10^2)\) and \(n_{\mathrm{test}}=O(10^4)\), the
\(n_{\mathrm{test}}k^2\) term dominates.

\subsection{Computing the predictive}
According to \citet{rasmussen2006gaussian}, the class probability for a test point is
\begin{equation}\label{predictive}
    p(\mathbf{y}^*|\mathbf{y},\mathbf{X}^*) = \mathbb{E}_{p(\mathbf{f}^*|\mathbf{y}, \mathbf{X}^*)}[\sigma(\mathbf{f}^*)] = \int\sigma(\mathbf{f}^*)p(\mathbf{f}^*|\mathbf{y}, \mathbf{X}^*)d\mathbf{f}^* 
\end{equation}

for $\mathbf{y}$ the training label, $\mathbf{X}^*$ the testing data, $\sigma$ the sigmoid function. The integral is approximated using $S$ posterior samples 
\citep{rasmussen2006gaussian}:
\begin{equation} \label{eq:mc_predictive}
    \hat{p}(y^* = 1 \mid \mathbf{y}, \mathbf{x}^*) 
    = \frac{1}{S} \sum_{s=1}^{S} \sigma\!\left(f_*^{(s)}\right),
\end{equation}
where $f_*^{(s)} \sim p(f^* \mid \mathbf{u}^{(s)}, \mathbf{x}^*)$ 
is a single draw from the DTC predictive conditional (Eq.~\ref{TestLatent}) 
and $\mathbf{u}^{(s)} = \mathbf{L}[\mathcal{P}, :] \, 
\boldsymbol{\nu}^{(s)}$ is constructed from the $s$-th HMC sample. 
Each term averages over both posterior uncertainty (via the HMC sample 
$\boldsymbol{\nu}^{(s)}$) and predictive uncertainty (via the 
conditional draw $f_*^{(s)}$). We use $S = 1000$ post-warmup samples.

\section{Experiments}

\subsection{Datasets and Feature Extraction}
We evaluate on two histopathology benchmarks for identifying breast cancer metastasis, \textbf{PCam} and \textbf{CAMELYON17-WILDS}. \textbf{PCam} (\citet{veeling2018rotation}; CC0~1.0 license) contains 327{,}680 RGB/color image patches ($96 \times 96$) divided into 262{,}144 training, 32{,}768 validation, and 32{,}768 test images, with binary labels indicating the presence of tumor tissue. \textbf{CAMELYON17-WILDS} (\citet{bandi2019camelyon17}; publicly available via the WILDS benchmark \citet{koh2021wilds}) is the \textbf{CAMELYON17} dataset accessed via the WILDS benchmark (\citet{koh2021wilds}). It contains 455{,}954 $96 \times 96$ patches divided into 302{,}000 training images from three hospitals, 34{,}000 valid images from the fourth hospital, and 85{,}000 testing images from the fifth hospital. Both are standard benchmarks for cancer diagnosis where calibrated uncertainty is relevant. In addition, $\textbf{CAMELYON17-WILDS}$ tests the method's calibration under distribution shift. We use the official \textbf{CAMELYON17-WILDS} splits for cross-hospital evaluation; because our classifiers operate on externally 
pretrained \textit{Phikon} representations, these results are not intended 
as a standard WILDS leaderboard submission.

\subsubsection*{Embedding as features}
The training datasets both have images of size 96 $\times$ 96 
pixels. Although ideally we can run GPC over raw pixels, the 
kernel matrices are not informative in such high dimensions. 
Thus, we use the 512-dimensional PCA projection of 
\textit{Phikon}'s \citep{filiot2023phikon} (Apache~2.0 license) 
768-dimensional CLS token output as features, where 
\textit{Phikon} is a ViT-Base transformer pretrained on 
$\sim$40M histopathology images. PCA was fitted on training-set 
embeddings only and applied unchanged to validation and test 
embeddings.

\subsection{Experiment 1: Methodological Validation}
Before scaling to full datasets, we validate our methodological choices 
on small-scale data ($N=2{,}000$). First, using synthetic two-class 
Gaussian blob data, we confirm that accelerated RPCholesky reduces 
kernel factorization time by 6$\times$ and total pipeline time by 
15$\times$ compared to dense factorization (averaged over 10 runs). 
Second, we compare three MCMC samplers --- Random-Walk Metropolis (RWM), 
the Metropolis-adjusted Langevin algorithm (MALA), and HMC --- on the 
low-rank GPC posterior with $k=200$. On synthetic data, MALA and HMC 
both outperform RWM in autocorrelation time and effective sample size 
per second, but do not clearly separate from each other. However, when 
repeating the comparison on a 2{,}000-point subsample of \textbf{PCam}, 
HMC establishes a clear advantage over both alternatives, motivating 
its use in subsequent full-scale experiments.

\subsection{Experiment 2: Full-scale HMC GPC}
We scale our method to the full training set of \textbf{PCam} and \textbf{CAMELYON17-WILDS} in both Experiment 2 and 3, while comparing against stochastic variational Gaussian process classification (SVGP) baseline in Experiment 2 and neural network baselines in Experiment 3. 

\paragraph{SVGP baseline.}
We use SVGP \citep{hensman2015scalable} as the primary scalable GP 
baseline, implemented in GPyTorch, trained with Adam 
(lr=$10^{-2}$, batch size 1024) for 50 epochs. The number of 
inducing points is fixed to $M=512$, with locations initialized 
from training embeddings and optimized jointly with the variational 
parameters and kernel hyperparameters (bandwidth initialized at 
$l=64$) during training. Each method is evaluated at its own best 
configuration.

\paragraph{Experiment setup.}
For our method, we factor the \(N \times N\) training kernel matrix 
using accelerated RPCholesky and sample the low-rank GP posterior 
using the non-centered parameterization in \cref{eq:NCP}. The HMC 
sampler uses leapfrog integration with step-size adaptation via dual 
averaging \citep{hoffman2014nuts}, following standard practice 
\citep{betancourt2017conceptual}, with \(n_{\mathrm{warmup}}=1000\) 
warmup steps and \(S=1000\) post-warmup samples. We select 
hyperparameters via sequential sweeps on a \(N=50{,}000\) subset of 
PCam (Appendix~\ref{app:sweeps}), fixing each before sweeping the 
next. Leapfrog steps 
(\(n_{\text{leapfrog}} \in \{10,16,25,35\}\)) are chosen first: all 
configurations produce identical predictive metrics, so we select 
\(n_{\text{leapfrog}}=25\) for low autocorrelation time and 
reasonable wall-clock cost. Rank (\(k \in \{100,200,300,500\}\)) is 
chosen next: performance peaks at \(k=200\) despite tighter kernel 
approximations at higher ranks, with ECE rising monotonically with 
rank. Bandwidth (\(l \in \{25,49,100,200\}\)) is chosen last: the 
approximate median pairwise distance yields the lowest NLL. We apply 
the same configuration to both datasets: \(n_{\text{leapfrog}}=25\), 
\(k=200\), and bandwidth set to the approximate median pairwise 
distance (\(l \approx 49\) on PCam, \(l \approx 54\) on 
CAMELYON17-WILDS). To avoid dataset-specific tuning, we reuse the same sampler and rank configuration on both datasets, while setting the RBF 
lengthscale to the training-set median pairwise distance.

\paragraph{Reproducibility.}
For SVGP and our method, we report mean and standard deviation over 
5 independent runs (seeds 1--5). For our method, each seed rebuilds 
the RPCholesky factorization (re-randomizing pivot selection), 
reinitializes HMC, and redraws all posterior and predictive samples. 
For SVGP, each seed reinitializes inducing locations, variational 
parameters, and mini-batch ordering. Neural baselines are reported 
from a single run due to longer training times; our primary 
comparison is between HMC and SVGP.

\subsection{Experiment 3: Neural Network Baselines}

We evaluate three neural network baselines on both datasets.

\begin{enumerate}
    \item \textbf{MLP:} a 2-layer MLP (512$\to$512$\to$1 with ReLU and 
dropout 0.3) trained on the same 512-dimensional Phikon embeddings 
used by our method, optimized with AdamW (lr=$10^{-3}$, weight 
decay $10^{-4}$) and cosine annealing for up to 50 epochs with 
early stopping (patience 5) based on validation AUROC;
\item \textbf{U-Net:} a U-Net encoder-decoder (base channels 32) 
that produces a dense logit map and spatially averages it for 
image-level classification, trained directly on raw $96\times96$ 
pixel inputs with Adam (lr=$10^{-3}$, weight decay $10^{-4}$) and 
class-balanced loss weighting, for 50 epochs on \textbf{PCam} and 
50 epochs on \textbf{CAMELYON17-WILDS}, selecting the best epoch 
by validation AUROC (best epoch 3 on both datasets);
\item \textbf{ViT-S:} a Vision Transformer-Small ($\sim$21.6M 
parameters, patch size 16, depth 12, 6 heads) trained from scratch 
on raw pixel inputs with lr=$3\times10^{-4}$, weight decay 0.05, 
batch size 32, for 50 epochs.
\end{enumerate}

\paragraph{Metrics} We report metrics covering classification performance, calibration, probabilistic predictive quality, and training time: accuracy, negative log-likelihood (NLL), expected calibration error (ECE), area under the receiver operating characteristic curve (AUROC), Brier score, and false negative rate (FNR). ECE is computed with 15 equal-width bins. FNR is computed using a decision threshold of 0.5.

\subsection{Results}
\paragraph{Computational Cost}
All of the experiments are run on a high-performance computing (HPC) cluster and on a single NVIDIA A100 GPU (80\,GB). Experiment 2 runs in approximately 22 minutes for \textbf{PCam}: Only 16 seconds for accelerated RPCholesky kernel factorization, and $\sim$20 minutes for HMC sampling and $\sim$2 seconds for predictive inference. By comparison, a dense Cholesky factorization for $N \approx 300,000$ requires 550 GB storage (for \textbf{PCam}, 
$N=262,144$, in float64) and is infeasible on a single 80 GB A100. 
Accelerated RPCholesky eliminates this bottleneck entirely: the factorization accounts for less than 2\% of total runtime, shifting the computational cost to HMC sampling. Our method, SVGP, and the MLP baseline operate on precomputed Phikon embeddings; the one-time embedding extraction cost is shared across these methods and excluded from all reported times. Among methods operating on the shared precomputed Phikon embeddings, our method requires approximately 22 minutes compared with approximately 5 minutes for SVGP. U-Net and ViT-S operate directly on raw images and include representation learning in their reported runtime, so these times are not directly comparable.

\paragraph{Data Analysis}
Among the neural baselines, U-Net achieves the best calibration-related metrics on both datasets. ViT-S underperforms due to its high data requirements when training from scratch without large-scale pretraining. The MLP underperforms relative to our method despite using the same Phikon embeddings, suggesting that the lightweight classification head lacks the capacity to fully exploit the embedding structure, whereas the GP kernel captures richer similarity information from the same representations.

Table~\ref{tab:results} reports results on both \textbf{PCam} 
and \textbf{CAMELYON17-WILDS}. On \textbf{PCam}, our method outperforms SVGP across all metrics (e.g., ECE 0.049 vs 0.087, accuracy 0.921 vs 0.905). 
On \textbf{CAMELYON17-WILDS}, where the test set comes from a 
held-out hospital, the gap widens dramatically: relative to SVGP, 
our method reduces ECE by 87.8\%, NLL by 43.3\%, Brier score 
by 76.7\%, and false negative rate by 89.8\%. Relative to U-Net (the best-calibrated neural baseline), it reduces ECE by 78.2\%, NLL 
by 65.4\%, Brier score by 83.6\%, and false negative rate by 
86.1\%.

\begin{figure}[H]
  \centering
  \includegraphics[width=0.48\linewidth]{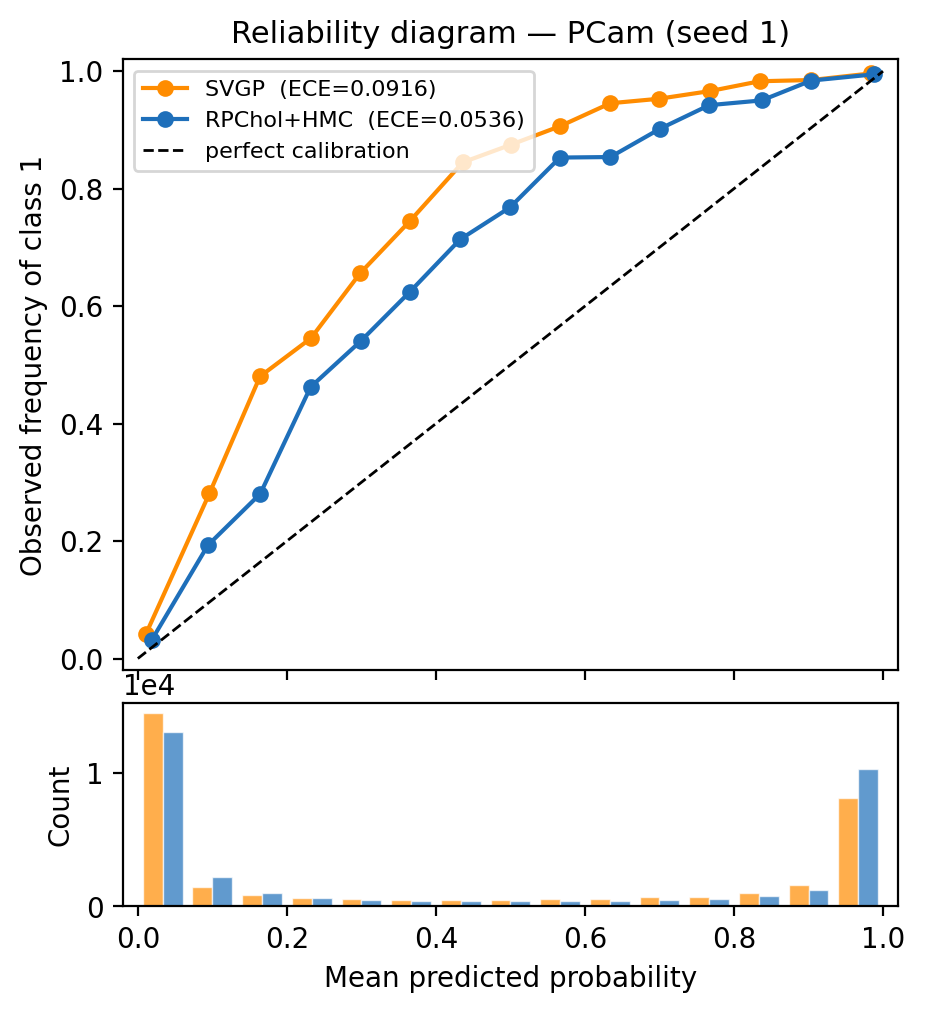}
  \hfill
  \includegraphics[width=0.48\linewidth]{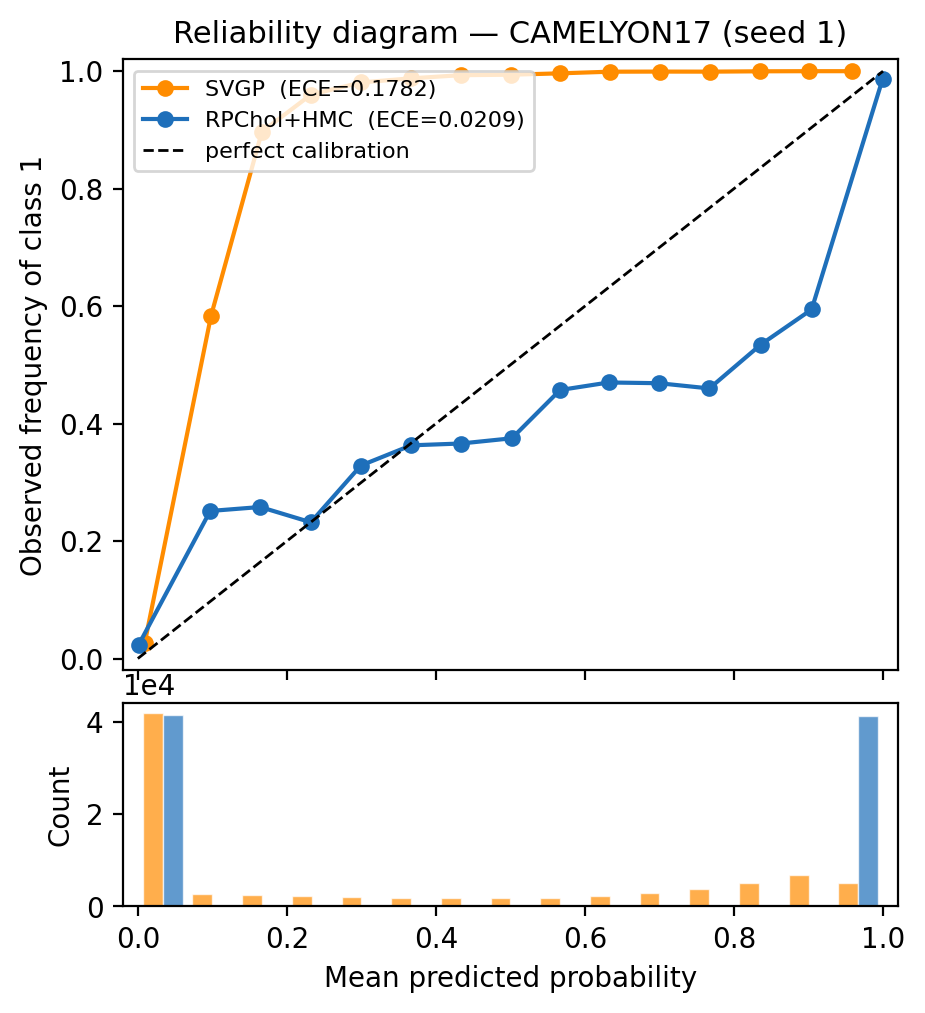}
  \caption{Reliability diagrams with prediction histograms on PCam 
  (left) and CAMELYON17-WILDS (right) for SVGP and our method 
  (seed 1, 15 equal-width bins). The dashed line indicates perfect 
  calibration. Under distribution shift (right), SVGP becomes 
  severely miscalibrated while our method is substantially closer 
  to the diagonal over the high-density probability regions 
  (bottom panels).}
  \label{fig:reliability}
\end{figure}

Figure~\ref{fig:reliability} shows reliability diagrams with 
prediction histograms illustrating the calibration difference: on 
CAMELYON17-WILDS, SVGP's predicted probabilities diverge sharply 
from observed frequencies, while our method is substantially 
closer to the diagonal over the high-density probability regions. 
The histograms show that both methods concentrate predictions near 
0 and 1, with few samples in intermediate bins where calibration 
curves may appear noisy.

\paragraph{Calibration on a held-out hospital.} 
\textbf{CAMELYON17-WILDS} evaluates generalization to a hospital not 
represented in the training set. On this held-out hospital, our 
method achieves an ECE of $0.021\pm0.001$, compared with 
$0.173\pm0.021$ for SVGP. The calibration advantage observed on 
\textbf{PCam} persists in the cross-hospital setting and becomes 
substantially larger relative to SVGP. Because \textbf{PCam} and 
\textbf{CAMELYON17-WILDS} are distinct datasets, we do not interpret the 
change in absolute ECE between the two benchmarks as a direct 
measure of degradation caused by distribution shift.

\begin{table}[H]
\centering
\resizebox{\linewidth}{!}{
\begin{tabular}{llcccccc}
\toprule
Dataset & Method
  & ECE $\downarrow$ & NLL $\downarrow$ & Brier $\downarrow$
  & FNR $\downarrow$ & Acc.\ $\uparrow$ & AUROC $\uparrow$ \\
\midrule
\multirow{5}{*}{\textbf{PCam}}
  & SVGP
    & .087{\scriptsize$\pm$.004}
    & .250{\scriptsize$\pm$.011}
    & .072{\scriptsize$\pm$.004}
    & .177{\scriptsize$\pm$.010}
    & .905{\scriptsize$\pm$.005}
    & .979{\scriptsize$\pm$.001} \\
  & U-Net
    & .059 & .401 & .110
    & .152 & .859 & .925 \\
  & MLP
    & .116 & .711 & .111
    & .234 & .877 & .965 \\
  & ViT-S
    & .191 & .814 & .206
    & .427 & .751 & .875 \\
  \cmidrule{2-8}
  & Low-rank HMC GP
    & \textbf{.049}{\scriptsize$\pm$.004}
    & \textbf{.200}{\scriptsize$\pm$.010}
    & \textbf{.059}{\scriptsize$\pm$.003}
    & \textbf{.137}{\scriptsize$\pm$.008}
    & \textbf{.921}{\scriptsize$\pm$.004}
    & \textbf{.980}{\scriptsize$\pm$.002} \\
\midrule
\multirow{5}{*}{\shortstack[l]{\textbf{CAM17}\\\textbf{-WILDS}}}
  & SVGP
    & .173{\scriptsize$\pm$.021}
    & .323{\scriptsize$\pm$.045}
    & .103{\scriptsize$\pm$.016}
    & .304{\scriptsize$\pm$.044}
    & .848{\scriptsize$\pm$.022}
    & \textbf{.994}{\scriptsize$\pm$.001} \\
  & U-Net
    & .092 & .505 & .142
    & .217 & .817 & .893 \\
  & MLP
    & .189 & 1.402 & .183
    & .377 & .811 & .956 \\
  & ViT-S
    & .387 & 2.078 & .397
    & .798 & .548 & .651 \\
  \cmidrule{2-8}
  & Low-rank HMC GP
    & \textbf{.021}{\scriptsize$\pm$.001}
    & \textbf{.183}{\scriptsize$\pm$.002}
    & \textbf{.024}{\scriptsize$\pm$.000}
    & \textbf{.031}{\scriptsize$\pm$.001}
    & \textbf{.973}{\scriptsize$\pm$.000}
    & .992{\scriptsize$\pm$.000} \\
\bottomrule
\end{tabular}
}
\caption{Predictive metrics on PCam (top) and CAMELYON17-WILDS (bottom).
SVGP and Low-rank HMC GP report mean over 5 seeds;
$\pm$ denotes one standard deviation.
Neural baselines are single-run.
$\downarrow$\,=\,lower is better,
$\uparrow$\,=\,higher is better.
Best value per dataset in \textbf{bold}.}
\label{tab:results}
\end{table}

\FloatBarrier
\section{Conclusion}
\subsubsection*{Discussion}
The most striking finding is that our method shows its largest calibration advantage on the held-out-hospital CAMELYON17-WILDS evaluation. While all methods perform reasonably on 
the in-distribution \textbf{PCam} benchmark, the gap widens substantially 
on \textbf{CAMELYON17-WILDS}, where test data comes from an unseen 
hospital. Variational inference based on the reverse KL objective has been shown to underestimate posterior variance \citep{blei2017variational,yao2018yes}, which may produce overconfident predictions that degrade when the test distribution differs from training. Our empirical results are consistent with this concern: on the held-out CAMELYON17-WILDS hospital, SVGP is substantially more miscalibrated than our HMC-based method. HMC-based sampling avoids imposing the Gaussian variational posterior used by SVGP and represents posterior uncertainty within the fixed low-rank model. This property is particularly relevant for clinical deployment, where a 
cancer detection model trained at one hospital must generalize reliably 
to others. In practice, well-calibrated predictive probabilities could, if validated at the slide and patient level, support institution-specific threshold selection that balances sensitivity and specificity according to local clinical priorities, rather than relying on a fixed threshold that may be miscalibrated at a new site.

Beyond calibration, we confirm that accelerated RPCholesky makes 
MCMC-based GPC feasible at modern dataset scale rather than purely 
theoretical. The factorization that was previously the computational 
bottleneck now accounts for less than 2\% of total runtime, shifting 
the cost to HMC warmup and sampling.

\subsubsection*{Limitations}
Our work has several limitations. First, we fix the kernel bandwidth 
$l$ via parameter sweep rather than sampling it jointly with the latent 
function $\mathbf{f}$. In a fully Bayesian treatment, $l$ would be 
inferred as part of the posterior, and fixing it may lead to suboptimal 
uncertainty estimates for some datasets. We additionally fix the kernel amplitude $s_f$ and do not evaluate sensitivity to this choice. Second, the sequential 
parameter sweep over leapfrog steps, rank, and bandwidth was guided in 
part by practitioner experience with MCMC tuning; a more systematic 
approach such as jointly optimizing these hyperparameters could yield 
improved results. Third, while we report mean and standard deviation over 5 seeds for our method and SVGP, the neural network baselines are single-run, and we do not perform formal statistical significance testing. Additionally, because our method and SVGP also differ in their inducing representations and kernel-hyperparameter treatment, the comparison does not isolate posterior inference as the sole cause of the observed calibration difference. Finally, our evaluation is performed on patch-level retrospective benchmarks and does not establish patient-level clinical utility.

\subsubsection*{Future Work}
The primary computational bottleneck is now HMC sampling, which takes 
approximately 20 minutes. Parallel chains, adaptive mass matrices, or 
the No-U-Turn Sampler \citep{hoffman2014nuts} could reduce this cost. 
More broadly, our framework operates on frozen embeddings from a 
histopathology foundation model, but the clinical workflow for breast 
cancer begins with mammographic screening, not biopsy. Extending this 
method to mammography datasets such as \textbf{EMBED} would require embeddings 
from models pretrained on radiological rather than histopathological 
images. If feasible, uncertainty-calibrated classification at the 
mammography stage could reduce unnecessary biopsies, providing clinical 
value earlier in the diagnostic pipeline.

\bibliographystyle{abbrvnat}
\bibliography{references}

@article{chen2025randomly,
  title         = {Randomly Pivoted Cholesky: Practical Approximation of a Kernel Matrix with Few Entry Evaluations},
  author        = {Chen, Yifan and Epperly, Ethan N. and Tropp, Joel A. and Webber, Robert J.},
  journal       = {Communications on Pure and Applied Mathematics},
  volume        = {78},
  number        = {5},
  pages         = {995--1041},
  year          = {2025},
  doi           = {10.1002/cpa.22234},
  eprint        = {2207.06503},
  archivePrefix = {arXiv},
  primaryClass  = {math.NA}
}

@article{epperly2025embrace,
  title         = {Embrace Rejection: Kernel Matrix Approximation by Accelerated Randomly Pivoted Cholesky},
  author        = {Epperly, Ethan N. and Tropp, Joel A. and Webber, Robert J.},
  journal       = {SIAM Journal on Matrix Analysis and Applications},
  year          = {2025},
  doi           = {10.1137/24M1699048},
  eprint        = {2410.03969},
  archivePrefix = {arXiv},
  primaryClass  = {math.NA}
}

@inproceedings{snelson2005sparse,
  title     = {Sparse Gaussian Processes using Pseudo-inputs},
  author    = {Snelson, Edward and Ghahramani, Zoubin},
  booktitle = {Advances in Neural Information Processing Systems},
  volume    = {18},
  year      = {2005},
  publisher = {MIT Press},
  url       = {https://papers.nips.cc/paper/2857-sparse-gaussian-processes-using-pseudo-inputs}
}

@inproceedings{williams2001nystrom,
  title     = {Using the {Nystr{\"o}m} Method to Speed Up Kernel Machines},
  author    = {Williams, Christopher K. I. and Seeger, Matthias},
  booktitle = {Advances in Neural Information Processing Systems 13},
  editor    = {Leen, Todd K. and Dietterich, Thomas G. and Tresp, Volker},
  pages     = {682--688},
  year      = {2001},
  publisher = {MIT Press},
  url       = {https://papers.nips.cc/paper/1866-using-the-nystrom-method-to-speed-up-kernel-machines}
}

@article{quinonero2005unifying,
  title   = {A Unifying View of Sparse Approximate Gaussian Process Regression},
  author  = {Qui{\~n}onero-Candela, Joaquin and Rasmussen, Carl Edward},
  journal = {Journal of Machine Learning Research},
  volume  = {6},
  number  = {65},
  pages   = {1939--1959},
  year    = {2005},
  url     = {https://jmlr.org/papers/v6/quinonero-candela05a.html}
}

@misc{betancourt2013hamiltonian,
  title         = {Hamiltonian Monte Carlo for Hierarchical Models},
  author        = {Betancourt, M. J. and Girolami, Mark},
  year          = {2013},
  eprint        = {1312.0906},
  archivePrefix = {arXiv},
  primaryClass  = {stat.ME},
  doi           = {10.48550/arXiv.1312.0906}
}

@article{nickisch2008approximations,
  title   = {Approximations for Binary Gaussian Process Classification},
  author  = {Nickisch, Hannes and Rasmussen, Carl Edward},
  journal = {Journal of Machine Learning Research},
  volume  = {9},
  number  = {67},
  pages   = {2035--2078},
  year    = {2008},
  url     = {https://jmlr.org/papers/v9/nickisch08a.html}
}

@inproceedings{veeling2018rotation,
  title         = {Rotation Equivariant CNNs for Digital Pathology},
  author        = {Veeling, Bastiaan S. and Linmans, Jasper and Winkens, Jim and Cohen, Taco and Welling, Max},
  booktitle     = {Medical Image Computing and Computer Assisted Intervention -- MICCAI 2018},
  series        = {Lecture Notes in Computer Science},
  volume        = {11071},
  pages         = {210--218},
  year          = {2018},
  publisher     = {Springer},
  doi           = {10.1007/978-3-030-00934-2_24},
  eprint        = {1806.03962},
  archivePrefix = {arXiv},
  primaryClass  = {cs.CV},
  url           = {https://arxiv.org/abs/1806.03962}
}

@article{bandi2019camelyon17,
  title   = {From Detection of Individual Metastases to Classification of Lymph Node Status at the Patient Level: The {CAMELYON17} Challenge},
  author  = {Bandi, Peter and Geessink, Oscar and Manson, Quirine and van Dijk, Marcory and Balkenhol, Maschenka and Hermsen, Meyke and Bejnordi, Babak Ehteshami and Lee, Byungjae and Paeng, Kyunghyun and Zhong, Aoxiao and Li, Quanzheng and Zanjani, Farhad Ghazvinian and Zinger, Svitlana and Fukuta, Keisuke and Komura, Daisuke and Ovtcharov, Vlado and Cheng, Shenghua and Zeng, Shaoqun and Thagaard, Jeppe and Dahl, Anders Bjorholm and Lin, Huangjing and Chen, Hao and Jacobsson, Ludwig and Hedlund, Martin and Cetin, Melih and Halici, Eren and Jackson, Hunter and Chen, Richard and Both, Fabian and Franke, J{\"o}rg and K{\"u}sters-Vandevelde, Heidi and Vreuls, Willem and Bult, Peter and van Ginneken, Bram and van der Laak, Jeroen A. W. M. and Litjens, Geert},
  journal = {IEEE Transactions on Medical Imaging},
  volume  = {38},
  number  = {2},
  pages   = {550--560},
  year    = {2019},
  doi     = {10.1109/TMI.2018.2867350}
}

@inproceedings{koh2021wilds,
  title     = {{WILDS}: A Benchmark of in-the-Wild Distribution Shifts},
  author    = {Koh, Pang Wei and Sagawa, Shiori and Marklund, Henrik and Xie, Sang Michael and Zhang, Marvin and Balsubramani, Akshay and Hu, Weihua and Yasunaga, Michihiro and Phillips, Richard Lanas and Gao, Irena and Lee, Tony and David, Etienne and Stavness, Ian and Guo, Wei and Earnshaw, Berton and Haque, Imran and Beery, Sara M. and Leskovec, Jure and Kundaje, Anshul and Pierson, Emma and Levine, Sergey and Finn, Chelsea and Liang, Percy},
  booktitle = {Proceedings of the 38th International Conference on Machine Learning},
  series    = {Proceedings of Machine Learning Research},
  volume    = {139},
  pages     = {5637--5664},
  year      = {2021},
  publisher = {PMLR},
  url       = {https://proceedings.mlr.press/v139/koh21a.html}
}

@article{filiot2023phikon,
  title        = {Scaling Self-Supervised Learning for Histopathology with Masked Image Modeling},
  author       = {Filiot, Alexandre and Ghermi, Ridouane and Olivier, Antoine and Jacob, Paul and Fidon, Lucas and Mac Kain, Alice and Saillard, Charlie and Schiratti, Jean-Baptiste},
  journal      = {medRxiv},
  year         = {2023},
  publisher    = {Cold Spring Harbor Laboratory Press},
  elocation-id = {2023.07.21.23292757},
  doi          = {10.1101/2023.07.21.23292757},
  url          = {https://www.medrxiv.org/content/early/2023/07/26/2023.07.21.23292757}
}

@inproceedings{titsias2009variational,
  title        = {Variational Learning of Inducing Variables in Sparse Gaussian Processes},
  author       = {Titsias, Michalis K.},
  booktitle    = {Proceedings of the Twelth International Conference on Artificial Intelligence and Statistics},
  pages        = {567--574},
  year         = {2009},
  organization = {PMLR}
}

@inproceedings{hensman2013gaussian,
  title     = {Gaussian Processes for Big Data},
  author    = {Hensman, James and Fusi, Nicolo and Lawrence, Neil D.},
  booktitle = {Proceedings of the Twenty-Ninth Conference on Uncertainty in Artificial Intelligence},
  pages     = {282--290},
  year      = {2013}
}

@inproceedings{hensman2015scalable,
  title        = {Scalable Variational Gaussian Process Classification},
  author       = {Hensman, James and Matthews, Alexander G. de G. and Ghahramani, Zoubin},
  booktitle    = {Proceedings of the Eighteenth International Conference on Artificial Intelligence and Statistics},
  pages        = {351--360},
  year         = {2015},
  organization = {PMLR}
}

@article{blei2017variational,
  title   = {Variational Inference: A Review for Statisticians},
  author  = {Blei, David M. and Kucukelbir, Alp and McAuliffe, Jon D.},
  journal = {Journal of the American Statistical Association},
  volume  = {112},
  number  = {518},
  pages   = {859--877},
  year    = {2017}
}

@inproceedings{yao2018yes,
  title        = {Yes, but Did It Work?: Evaluating Variational Inference},
  author       = {Yao, Yuling and Vehtari, Aki and Simpson, Daniel and Gelman, Andrew},
  booktitle    = {Proceedings of the 35th International Conference on Machine Learning},
  pages        = {5581--5590},
  year         = {2018},
  organization = {PMLR}
}

@inproceedings{guo2017calibration,
  title        = {On Calibration of Modern Neural Networks},
  author       = {Guo, Chuan and Pleiss, Geoff and Sun, Yu and Weinberger, Kilian Q.},
  booktitle    = {Proceedings of the 34th International Conference on Machine Learning},
  pages        = {1321--1330},
  year         = {2017},
  organization = {PMLR}
}

@inproceedings{ovadia2019trust,
  title     = {Can You Trust Your Model's Uncertainty? Evaluating Predictive Uncertainty Under Dataset Shift},
  author    = {Ovadia, Yaniv and Fertig, Emily and Ren, Jie and Nado, Zachary and Sculley, D. and Nowozin, Sebastian and Dillon, Joshua V. and Lakshminarayanan, Balaji and Snoek, Jasper},
  booktitle = {Advances in Neural Information Processing Systems},
  volume    = {32},
  year      = {2019}
}

@inproceedings{ronneberger2015unet,
  title     = {{U-Net}: Convolutional networks for biomedical image segmentation},
  author    = {Ronneberger, Olaf and Fischer, Philipp and Brox, Thomas},
  booktitle = {MICCAI},
  year      = {2015}
}

@inproceedings{dosovitskiy2021vit,
  title     = {An image is worth {16x16} words: Transformers for image recognition at scale},
  author    = {Dosovitskiy, Alexey and others},
  booktitle = {ICLR},
  year      = {2021}
}

@article{hoffman2014nuts,
  title   = {The No-U-Turn Sampler: Adaptively Setting Path Lengths in Hamiltonian Monte Carlo},
  author  = {Hoffman, Matthew D. and Gelman, Andrew},
  journal = {Journal of Machine Learning Research},
  volume  = {15},
  number  = {47},
  pages   = {1593--1623},
  year    = {2014},
  url     = {https://jmlr.org/papers/v15/hoffman14a.html}
}

@article{betancourt2017conceptual,
  title   = {A Conceptual Introduction to Hamiltonian Monte Carlo},
  author  = {Betancourt, Michael},
  journal = {arXiv preprint arXiv:1701.02434},
  year    = {2017},
  url     = {https://arxiv.org/abs/1701.02434}
}

@book{rasmussen2006gaussian,
  title     = {Gaussian Processes for Machine Learning},
  author    = {Rasmussen, Carl Edward and Williams, Christopher K. I.},
  year      = {2006},
  publisher = {MIT Press},
  address   = {Cambridge, MA},
  url       = {https://gaussianprocess.org/gpml/}
}

@incollection{neal2011mcmc,
  title     = {{MCMC} Using {Hamiltonian} Dynamics},
  author    = {Neal, Radford M.},
  booktitle = {Handbook of {Markov} Chain {Monte} {Carlo}},
  editor    = {Brooks, Steve and Gelman, Andrew and Jones, Galin L. and Meng, Xiao-Li},
  chapter   = {5},
  pages     = {113--162},
  year      = {2011},
  publisher = {Chapman and Hall/CRC},
  address   = {Boca Raton, FL}
}

@article{cooper2025modernizing,
  title   = {Modernizing Full Posterior Inference for Surrogate 
             Modeling of Categorical-Output Simulation Experiments},
  author  = {Cooper, Andrew and Booth, Annie S. and Gramacy, Robert B.},
  journal = {arXiv preprint arXiv:2501.14946},
  year    = {2025}
}


\appendix

\section{Parameter Sweep Details}
\label{app:sweeps}

\paragraph{Sweep 1: Leapfrog Steps.}
Configuration: $N = 50{,}000$, $k = 200$, $n_{\text{warmup}} = 1000$, 
$n_{\text{samples}} = 1000$, bandwidth set to approximate median 
pairwise distance. We report integrated autocorrelation time for the 
latent variables ($\tau_{\nu}$) and log-posterior ($\tau_{\log p}$).

\begin{center}
\begin{tabular}{c c c c c c}
  \toprule
  $n_{\text{leapfrog}}$ & $\tau_\nu$ & $\tau_{\log p}$ 
    & Accept rate & Final step size & Wall time (s) \\
  \midrule
  10 & 21.70 & 11.78 & 0.895 & 0.0335 & 1769.6 \\
  16 & 8.43  & 3.98  & 0.953 & 0.0343 & 573.4 \\
  25 & 4.06  & 2.58  & 0.810 & 0.0340 & 873.6 \\
  35 & 1.80  & 2.53  & 0.854 & 0.0340 & 2788.2 \\
  \bottomrule
\end{tabular}
\end{center}

\paragraph{Sweep 2: Rank.}
Configuration: $N = 50{,}000$, $n_{\text{leapfrog}} = 25$, 
$n_{\text{warmup}} = 750$, $n_{\text{samples}} = 500$, 
bandwidth = approx.\ median. Trace error is the relative trace-norm 
error $\|\mathbf{K} - \mathbf{L}\mathbf{L}^T\|_{\mathrm{tr}} 
/ \|\mathbf{K}\|_{\mathrm{tr}}$.

\begin{center}
\begin{tabular}{c c c c c c}
  \toprule
  Rank $k$ & Trace error & NLL & ECE & AUROC & Accuracy \\
  \midrule
  100 & 14.2\% & 0.196 & 0.0445 & 0.980 & 0.921 \\
  200 & 10.8\% & \textbf{0.186} & 0.0464 & \textbf{0.983} 
      & \textbf{0.927} \\
  300 & 8.6\%  & 0.192 & 0.0509 & 0.982 & 0.924 \\
  500 & 6.7\%  & 0.194 & 0.0525 & 0.983 & 0.923 \\
  \bottomrule
\end{tabular}
\end{center}

\paragraph{Sweep 3: Bandwidth.}
Configuration: $N = 50{,}000$, $k = 200$, $n_{\text{leapfrog}} = 25$, 
$n_{\text{warmup}} = 750$, $n_{\text{samples}} = 500$.

\begin{center}
\begin{tabular}{c c c c c c}
  \toprule
  Bandwidth & Trace error & NLL & ECE & AUROC & Accuracy \\
  \midrule
  25  & 48.3\% & 0.195 & 0.0500 & 0.982 & \textbf{0.929} \\
  49  & 11.1\% & \textbf{0.192} & 0.0504 & 0.982 & 0.923 \\
  100 & 1.8\%  & 0.195 & 0.0536 & 0.982 & 0.924 \\
  200 & 0.4\%  & 0.216 & 0.0718 & 0.981 & 0.923 \\
  \bottomrule
\end{tabular}
\end{center}

\newpage

\end{document}